\documentclass[journal]{IEEEtran}
\usepackage{amsmath}
\usepackage{graphicx}
\ifCLASSINFOpdf
\else
\fi
\begin{document}
\title{Improving Color Constancy by Discounting the Variation of Camera Spectral Sensitivity}

\author{Shao-Bing Gao, Ming Zhang, Chao-Yi Li, and Yong-Jie Li,~\IEEEmembership{Member,~IEEE}
% <-this % stops a space  Xian-Shi Zhang,
\thanks{Manuscript received September 21, 2015; This work was supported by the Major State Basic Research
Program (\#2013CB329401) and the Natural Science Foundations of China (\#91420105, \#61375115).
The work was also partially supported by the 111 Project of China under Grant B12027.
(Corresponding author: Yong-Jie Li)}% <-this % stops a space
\thanks{S.-B. Gao, M. Zhang, and Y.-J. Li are with the School of Life
Science and Technology, University of Electronic Science and Technology of
China, Chengdu 610054, China (email: \{gao\_shaobing, zm\_uestc\}@163.com, liyj@uestc.edu.cn).}% <-this % stops a space
\thanks{C.-Y. Li is with the School of Life Science and Technology, University of Electronic
Science and Technology of China, Chengdu 610054, China, and the Center for
Life Sciences, Shanghai Institutes for Biological Sciences, Chinese Academy
of Sciences, Shanghai 200031, China (email:cyli@sibs.ac.cn)}}
\maketitle

\begin{abstract}
%Computational color constancy (CC) has been specially designed to recover the true color of scene by first inferring the light source color and then discounting it from the captured color biased image.
%The image colors are determined not only by the scene illuminant but also by the camera spectral sensitivity (CSS). This paper studies the CSS effects on illuminant estimation arising in inter-dataset setup, namely the inter-dataset-based color constancy (inter-CC). We show the deficiency of existing learning-based CC algorithms when dealing with this inter-CC problem, i.e., training a CC model on one dataset and then testing it on another dataset that was captured by a distinct CSS. This is a very challengeable task that limits the wide application of many existing CC models. To overcome this inter-CC problem, we propose a simple way to solve it by firstly learning a transform matrix that converts the responses to the illuminant and images under one CSS into those under another CSS. Then, the learned matrix is used to adapt the data rendered under one CSS into another specific CSS before applying the CC model on another dataset. Theoretical analysis and experimental validation on synthetic, hyperspectral, and real camera captured images show that the proposed method can significantly improve the inter-CC performance for traditional CC algorithms. Moreover, by modeling those effects, we also get a truly color constancy image, which is invariant to changes of illuminant and camera sensors.
It is an ill-posed problem to recover the true scene colors from a color biased image by discounting the effects of scene illuminant and camera spectral sensitivity (CSS) at the same time. Most color constancy (CC) models have been designed to first estimate the illuminant color, which is then removed from the color biased image to obtain an image taken under white light, without the explicit consideration of CSS effect on CC. This paper first studies the CSS effect on illuminant estimation arising in the inter-dataset-based CC (inter-CC), i.e., training a CC model on one dataset and then testing on another dataset captured by a distinct CSS. We show the clear degradation of existing CC models for inter-CC application. Then a simple way is proposed to overcome such degradation by first learning quickly a transform matrix between the two distinct CSSs (CSS-1 and CSS-2). The learned matrix is then used to convert the data (including the illuminant ground truth and the color biased images) rendered under CSS-1 into CSS-2, and then train and apply the CC model on the color biased images under CSS-2, without the need of burdensome acquiring of training set under CSS-2. Extensive experiments on synthetic and real images show that our method can clearly improve the inter-CC performance for traditional CC algorithms. We suggest that by taking the CSS effect into account, it is more likely to obtain the truly color constant images invariant to the changes of both illuminant and camera sensors.
%Thus, we suggest that for better generalization of the state-of-the-art CC algorithms, the effect of CSS should be reasonably considered.
\end{abstract}

\begin{IEEEkeywords}
color constancy, camera spectral sensitivity, illuminant estimation, color correction.
\end{IEEEkeywords}

% \fi
\IEEEpeerreviewmaketitle

\section{Introduction}
\begin{figure}[t]
\begin{center}
\includegraphics[angle=0,width=0.42\textwidth]{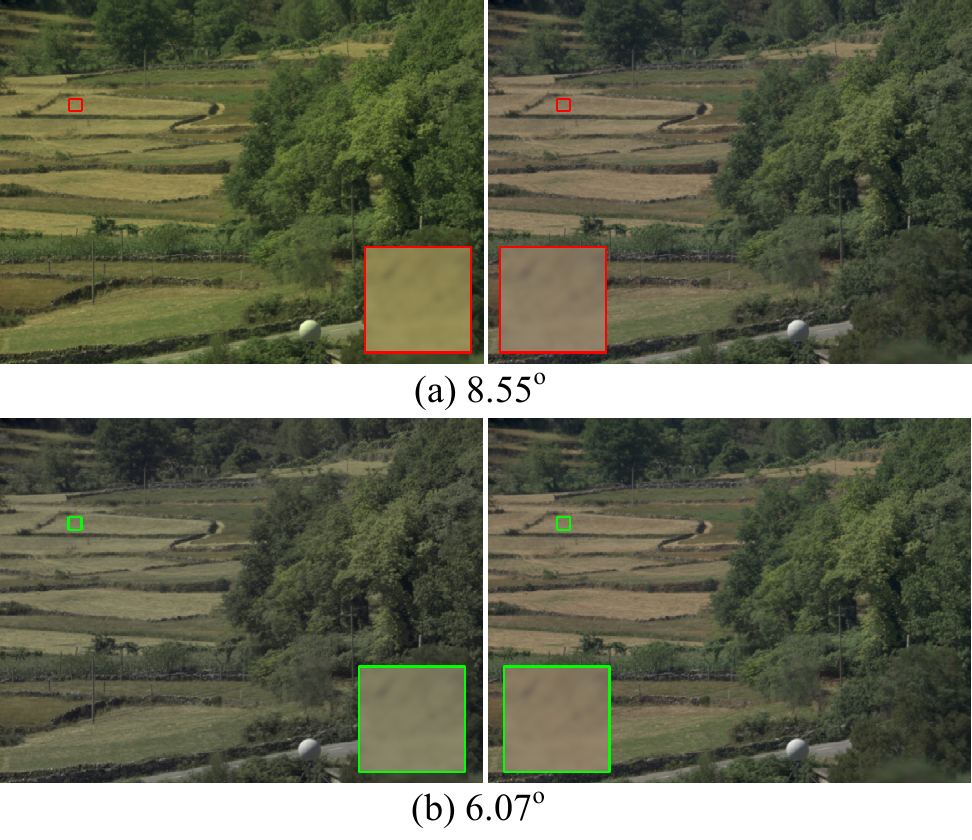}
\end{center}
\vspace{-1em}
\caption{The effects of camera spectral sensitivity (CSS) and
illuminant on image color appearance. (a) Images rendered with same CSS but
under different illuminants (left: green light source, right: white light source).
(b) Images rendered with different CSSs but under same illuminant (both are white light source).
Chromatic angular error values between the corresponding patches are given at the bottom of each row.}
\label{fig1}
\end{figure}

\IEEEPARstart{W}{ith} the rapid proliferation of digital imaging and videoing, accurate recording of the true color of a scene through the device-captured image is of extreme importance for many practical applications, ranging from the color-based object recognition and tracking to the quality control of textile and food processing \cite{vrhel2005color,funt1998machine,gijsenij2011computational}. However, these device-captured image colors are always affected by the prevailing changed light source color incident in a scene. Fig. \ref{fig1}(a) indicates that the images of a same scene rendered under two different illuminants obviously exhibit various color appearance. Thus, for the sake of maintaining the true color appearance of objects, the color artifacts due to the illuminant should be carefully eliminated.

Color constancy (CC) refers to the perceptual constancy of the human visual
system, which enables the perceived color of objects in a scene largely constant as the light source color changes
\cite{gijsenij2011computational,foster2011color,gao2013color}.
Many CC algorithms have been specially designed to imitate this visual attribute by computationally
estimating the illuminant and then removing the color cast to discount the bias due to the
illuminant (see \cite{gijsenij2011computational,foster2011color,hordley2006scene}
for excellent reviews). Recently, the performance of CC on several benchmark datasets has been
significantly progressed, especially for the state-of-the-art CC models that are
based on extensive feature extraction and machine learning techniques
\cite{cheng2015effective,finlayson2013corrected,joze2014exemplar,gijsenij2011color,
bianco2014adaptive,bianco2015color,li2015multi}.

However, as indicated in Fig. \ref{fig1}, image color is not only influenced by the scene light source color.
Actually, during the image acquisition phase, there are three factors influencing
the color values that we finally measure at each pixel, i.e., the reflectance of the objects
in the scene, the illuminant incident in a scene, and the camera spectral sensitivity (CSS).
While CC always devotes to obtain a stable color representation of scene across the changes of the illuminant
(Fig. \ref{fig1}(a)), the CSS also affects the color appearance of the scene (Fig. \ref{fig1}(b)).
\begin{figure}[t]
\begin{center}
\includegraphics[angle=0,width=0.5\textwidth]{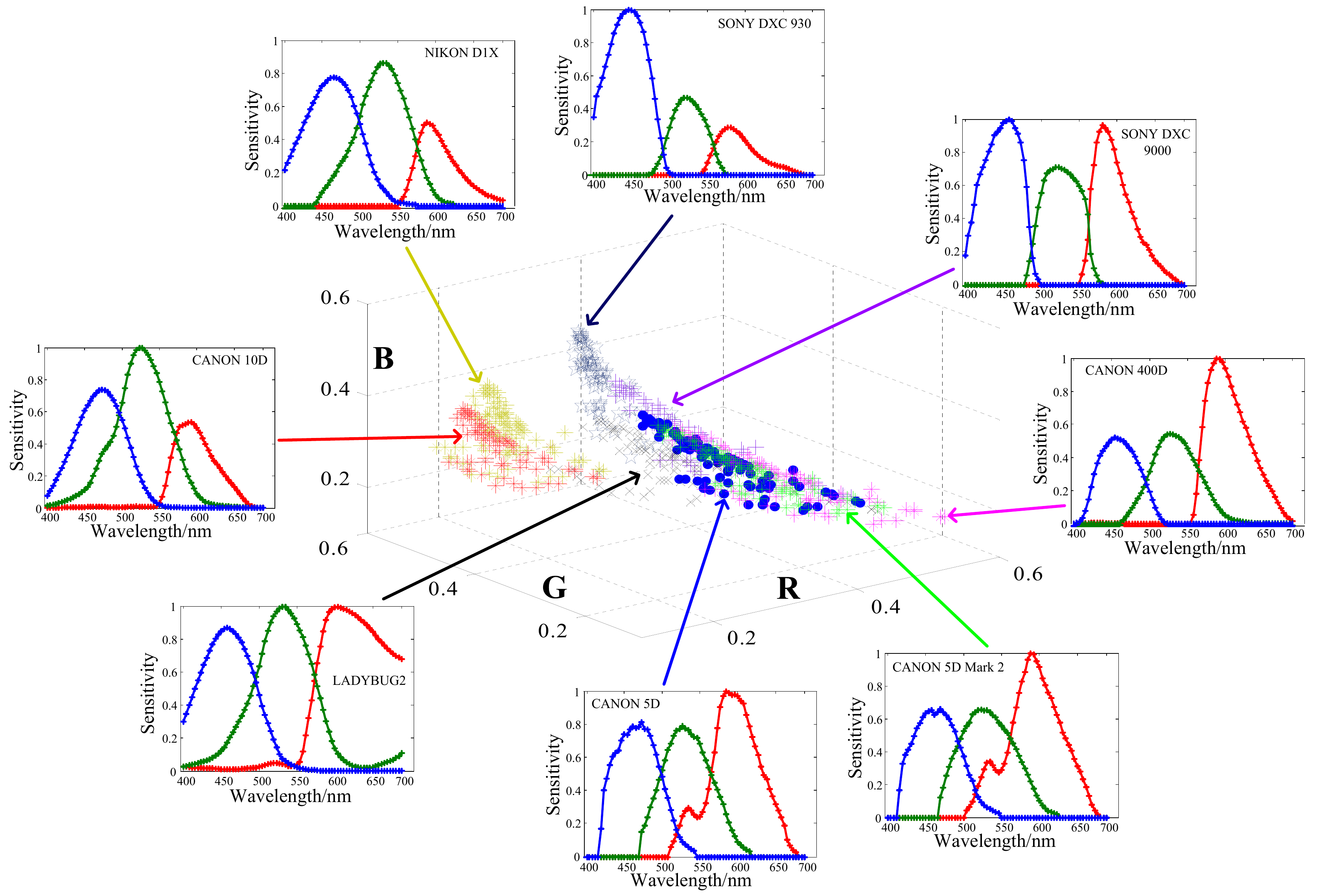}
\end{center}
\vspace{-1em}
\caption{Chromatic distributions of light source spectra \cite{barnard2002data}
responding to various CSSs \cite{zhao2009estimating,kawakami2013camera}. Each chromaticity
represents one color domain of 108 spectra rendered under a CSS.}
\label{fig2}
\end{figure}

So far, however, almost all existing benchmark datasets were collected using one camera with fixed CSS \cite{cheng2014illuminant,
barnard2002data,ciurea2003large,gehler2008bayesian,funt2010rehabilitation,vazquez2009color}
and many state-of-the-art learning-based CC algorithms implicitly assume that the images in a dataset are
captured by the same CSS, and are limited to evaluate their performance on intra-dataset-based CC (intra-CC)
\cite{gijsenij2011computational,cheng2015effective,finlayson2013corrected,gijsenij2011color,bianco2014adaptive,
bianco2015color,li2015multi,cardei2002estimating,funt2004estimating,shi2011illumination},
i.e., learning the model on one part of the dataset and testing the learned model on another part of the dataset.
Thus, the effect of CSS on CC can not be deeply probed according to the intra-CC
strategy. However, as indicated in
Fig. \ref{fig1}(b) and Fig. \ref{fig2}, both the color distributions of
the recorded images and the extracted true illuminants in the datasets rely on CSS.

While an existing intra-CC strategy discounts the color bias induced by the illuminant, the recovered image color
is in fact the appearance of the combination of the object reflectance and the camera sensor effect (Fig. \ref{fig1}(b)).
As a consequence, the existing CC algorithms may suffer problems when
dealing with the inter-dataset-based (inter-CC) application, i.e., training a model on one dataset
that was captured by a specific camera and then testing the learned model on another dataset
that was captured by another camera with different CSS.

In this paper, we draw insights into the rich literature on intrinsic image research,
which aims to decompose an image into various individual physical characteristics (e.g.,
reflectance and shade) that are independent of both illuminant and camera sensor
\cite{serra2014photometry,barrowrecovering,barron2015shape}.
To this end, we primarily focus on studying the effect of the CSS on CC arising in the inter-CC setup.

Specifically, we first point out that the chromatic distributions of both
the measured illuminants and recorded images by various CSSs are quite different,
even for a same scene under one illuminant, which is one of the significant causes
that results in the failure of inter-CC evaluation using current learning-based
CC algorithms. Then, in order to overcome this drawback, we propose a simple yet efficient
framework that incorporates the information of CSS into the process of CC. This is the main contribution of this work. In particular, we first
learn a transform matrix between the CSS functions of two distinct cameras (CSS-1 and CSS-2). Then, the learned matrix is used to convert the color biased images and the provided
illuminants recorded with CSS-1 into those with CSS-2 before training the model and testing on the image(s)
recorded with CSS-2.

%This simple strategy could be treated as a kind of transfer learning application in CC by using the prior of CSS, which
%can significantly enhance the generalization of learning-based
%CC algorithms when training them on one dataset and applying them on another dataset captured by a distinct
%CSS.
Moreover, by taking into account the CSS information, we also demonstrate how to obtain a stable color image representation of the scene that is almost independent of both illuminant and camera sensor. This stable color image representation may benefit the further computer vision applications such as intrinsic image decomposition, 3-D view synthesis, physics-based reflectance descriptor, and so on.

Although it is well-known that the CSS affects the image formation, how to discount such a prevail adverse effect in an elegant way and thus improve the performance of other color applications (e.g., CC) is a very difficult problem. In fact, some CC researches have also been devoted to the effect of imaging sensors on chromatic adaptation, in which both the source and destination illuminants are known. For example, the spectral sharpening of sensors attempts to simplify the illuminant change characterization and therefore improve the performance of any CC algorithm that is based on the diagonal-matrix transformation \cite{finlayson1994color,finlayson1994spectral}.

Totally different from those attempts, our aims are to probe into the possible effects
of various CSSs on inter-CC performance for traditional learning-based CC
models and to propose feasible solution to solve such a challenging problem. To the best of our knowledge, such issue has not been explicitly studied before in the area of CC. The motivation is that though learning-based models work well for intra-CC application, they require the collection of extensive training set for each CSS. In contrast, it would be undoubtedly more practically valuable if we could always obtain inter-CC performance competitive to that by learning-based intra-CC for any CSS, simply based on the collection of training set for only one CSS.

The rest of this paper is organized as follows. Section \uppercase\expandafter{\romannumeral2}
formulates the problem. The proposed solution is described in section
\uppercase\expandafter{\romannumeral3}. Section \uppercase\expandafter{\romannumeral4} presents the experiments to
validate our theoretical analysis and the proposed method. We conclude in Section \uppercase\expandafter{\romannumeral5}
by discussing some contributions and limitations of this work.
%\hfill September 17, 2014
%\cite{gijsenij2011computational}.
\begin{figure}[t]
\begin{center}
\includegraphics[angle=0,width=0.5\textwidth]{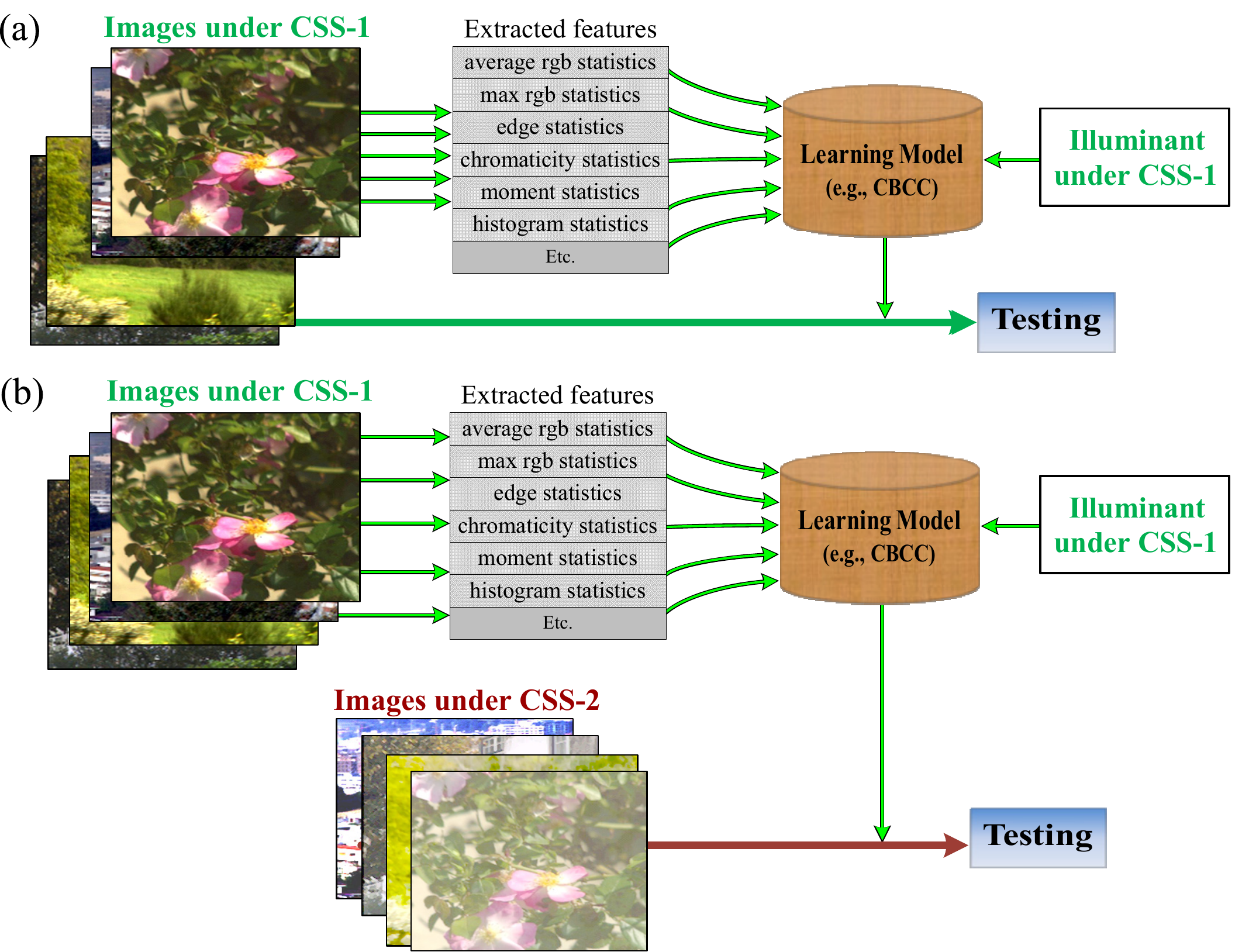}
\end{center}
\vspace{-1em}
\caption{The general framework of traditional color constancy (CC) evaluation. (a) intra-dataset-based CC, (b) inter-dataset-based CC.}
\label{fig3}
\end{figure}

\section{Color image fundamentals and color constancy}
As did in many papers, in this work we reasonably assume a single light source color across a scene
\cite{gijsenij2011computational,
cheng2015effective,gao2014efficient,gao2015color,yang2015efficient}.
Based on the common form of the linear imaging equation, the interaction of surface, light source,
and sensor can be indicated by a simple equation written as \cite{gijsenij2011computational,foster2011color}
\begin{equation}
\begin{aligned}
f_{c}(x)\!\!=\!\!\!\int_{\omega}S_{c}(\lambda)I(\lambda)C(x,\lambda)d\lambda
\end{aligned}
\vspace{-0.3em}
\end{equation}
where the integral is taken over the visible spectrum ${\omega}$ and $c\!\in\!\{R,G,B\}$ are
sensor channels. Equation (1) states that the captured image values $f(x)\!\!=\!\![f_{R}(x),f_{G}(x),f_{B}(x)]^{T}$
directly depend on the color of the light source $I(\lambda)$, the surface reflectance $C(x,\lambda)$ and
the camera spectral sensitivity $S(\lambda)\!\!=\!\![S_R(\lambda), S_G(\lambda), S_B(\lambda)]$, where $x$ is the spatial coordinate
and $\lambda$ is the wavelength of the light.

Earlier researches have demonstrated that it is sufficient to simulate the color transform of image
induced by the illuminant with a diagonal transform \cite{gijsenij2011computational,hordley2006scene,
ebner2009color,finlayson2001color,finlayson1994color}.
Thus, based on the diagonal transform assumption in CC, Eq (1) can be simplified as
\begin{equation}
\begin{aligned}
f_{c}(x)=I_{c}C_{c}(x)
\end{aligned}
\vspace{-0.3em}
\end{equation}
In general, CC aims at removing the color bias in captured image $f_{c}(x)$ by first estimating
the illuminant $I_{c}$, $c\!\in\!\{R,G,B\}$, then recovering $C_{c}(x)$ by dividing $f_{c}(x)$ by $I_{c}$.
However, both $I_{c}$ and $C_{c}(x)$ in Eq (2) are usually unknown and hence, given only the image
values $f_{c}(x)$, estimating $I_c$ is a typical ill-posed problem that cannot be solved without further
constraints \cite{gao2014efficient,buchsbaum1980spatial,land1971lightness,
van2007edge,finlayson2004shades}.
%\IEEEpubidadjcol

While many existing algorithms rely on aforementioned steps to achieve CC, it
should be explicitly pointed out that these algorithms are camera dependent (e.g., a fixed camera
\cite{finlayson2013corrected}).
Unfortunately, almost all current CC models only focus on developing techniques to eliminate
the color bias in image induced by the illuminant but ignoring the fact that the CSS
also contributes to the color bias when recording the images. In particular, for almost
all the color constancy datasets, the illuminant ground truth of each image is extracted
from a specific local region (e.g., the grey ball or color-checker) of the image recorded
by the CSS. That is, the illuminant ground truth is also CSS dependant.

For a clear illustration, we express the sensor responses to both the illuminant and surface respectively as
\cite{finlayson2013corrected}
\begin{equation}
\begin{aligned}
I_{c}\!\!=\!\!\!\int_{\omega}\!\!S_{c}(\lambda)I(\lambda)d\lambda
\end{aligned}
\vspace{-0.3em}
\end{equation}
\begin{equation}
\begin{aligned}
C_{c}(x)\!\!=\!\!\!\int_{\omega}\!\!S_{c}(\lambda)C(x,\lambda)d\lambda
\end{aligned}
\vspace{-0.3em}
\end{equation}
Obviously, both $I_{c}$ and $C_{c}(x)$ depend on the camera spectral sensitivity $S_{c}(\lambda)$.
Fig. \ref{fig1}(b) intuitively shows that the same scene rendered under
white light source but exhibit obvious color difference since two images ($C_{c}(x)$) are respectively
captured by two cameras with different CSSs. Similarly, the responses to illuminants ($I_{c}$) under
diverse CSSs also visually display in various color domains (Fig. \ref{fig2}).
\subsection{About the learning-based CC}
We abstract the standard framework of traditional learning-based CC as
\begin{equation}
\begin{aligned}
I_{c}\!\approx \!\Gamma\{K(f_{c}(x))\}
\end{aligned}
\vspace{-0.3em}
\end{equation}
where $I_{c}$ indicates a set of vectors of ground truth illuminants supplied by the dataset.
$K(f_{c}(x))$ represents a set of vectors of certain statistics extracted from the input image $f_{c}(x)$. $\Gamma$
donates a certain model that is committed to learn a transformation, which can effectively map the extracted features of images
$K(f_{c}(x))$ to the corresponding illuminant ground truth $I_{c}$. Basically, for the sake of comprehensive consideration
of the effectiveness and efficiency, various algorithms train their models using different machine learning techniques
$\Gamma$ and feature descriptors $K(f_{c}(x))$ for CC.

For example, the classical gamut mapping method trains a model that is able to correlate the gamut of the input
image to the canonical gamut under white light source of a fixed camera
\cite{forsyth1990novel,gijsenij2010generalized,finlayson2001color1}.
While statistical methods relate the CC as one kind of parameter
inference problem by assuming that the reflectance and illuminant meet specific
probability distribution \cite{gehler2008bayesian,brainard1997bayesian,chakrabarti2012color,rosenberg2003bayesian}, they need to train a model that can capture the mapping between the preprocessed images and the illuminant ground truth.
\begin{figure*}[t]
\begin{center}
\includegraphics[angle=0,width=0.8\textwidth]{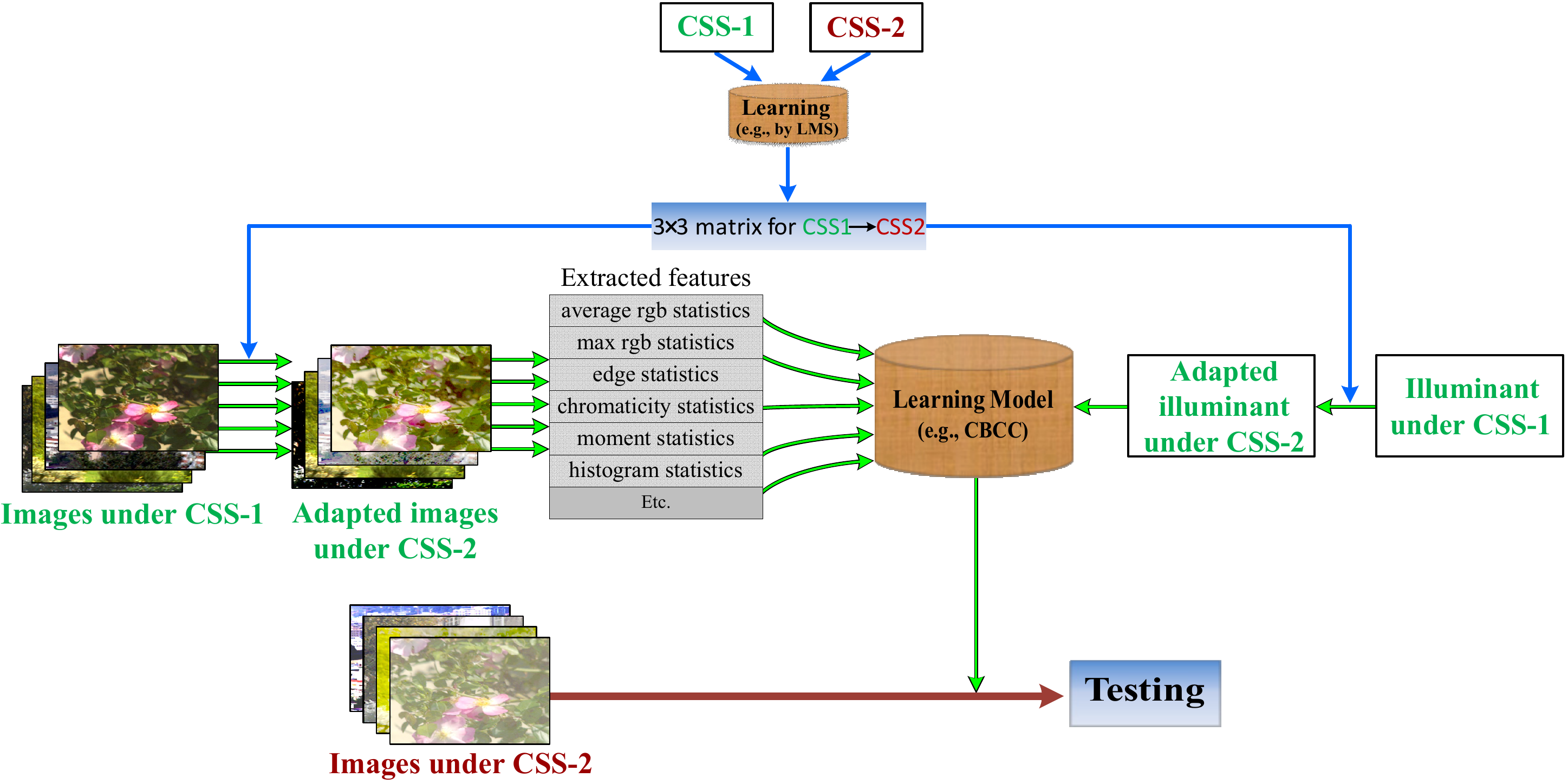}
\end{center}
\vspace{-1em}
\caption{The general framework of the proposed inter-dataset-based CC improved by taking
the CSS difference between datasets into account during the CC model training.}
\label{fig5}
\end{figure*}

A large class of learning-based methods inherently treat CC as a regression problem
\cite{finlayson2013corrected,funt2004estimating,
agarwal2006estimating}. This kind of methods could be unified under a standard regressive equation like
$I_{c}\!\!\approx \!\!K(f_{c}(x))L$. Here $L$ donates a regression matrix that needs
to be learned. The difference among regression based models is reflected in the specific choice of the
technique of learning a regression matrix $L$ and the selection of the features $K(f_{c}(x))$
adopted by each CC algorithm. Earlier approaches use nonlinear neural networks to learn a regression
matrix that can map the binarized chromaticity histogram of the input image to the illuminant \cite{cardei2002estimating}.
Similar approaches apply support vector regression [21], linear regression \cite{agarwal2006estimating,agarwal2009illumination},
or thin-plate spline interpolation techniques \cite{shi2011illumination} to learn the regression matrix using
almost the similar type of input data. More recently, the leading performance is obtained by employing more effective features
(color-edge moment \cite{finlayson2013corrected}), more efficient regression techniques
(e.g., regression trees \cite{cheng2015effective}), or even deep learning \cite{bianco2015color}.

Alternatively, if we replace the features $K(f_{c}(x))$ with the illuminant estimated by
multiple low-level based CC algorithms
\cite{gao2013color,gao2014efficient,buchsbaum1980spatial,land1971lightness,
van2007edge,finlayson2004shades},
the regression problem mentioned above is boiled down to the so-called committed-based CC \cite{bianco2010automatic,
schaefer2005combined,cardei1999committee,bianco2008consensus}. In this protocol,
$L$ is no longer regarded as a feature mapping matrix but a weighted matrix
that tries to 'optimally' fuse the output of multiple CC methods as a single
illuminant estimate under certain rule (e.g., the weights are optimized in the
least mean square (LMS) sense). Or in \cite{joze2014exemplar,gijsenij2011color,
bianco2008improving,van2007using,sapiro1999color}, $L$ is taken as an vote matrix that can select the most
appropriate CC method or previously stored illuminant for every input image according to the intrinsic
properties of natural images, e.g., high level information \cite{gijsenij2011color,bianco2010automatic,bianco2008improving,
van2007using,lu2009color,vazquez2012color}.
\subsection{Intra-dataset-based CC (Intra-CC)}
The common way to benchmark
the performance of a learning-based CC algorithm is to adopt the intra-dataset-based
evaluation with the form of so-called $n$-folds cross validation \cite{gijsenij2011computational,finlayson2013corrected,
li2015multi,gao2015color}.
For example, the dataset is first divided into $n$ parts ($n$-folds). Next, by applying the model
on the $n\!-\!1$ parts, the optimal parameters and structures of the model are trained using the
corresponding illuminant ground truth. Then, the trained model is tested on the remaining one
part of the data. For a complete procedure of $n$-folds cross validation, the steps mentioned above
are repeated $n$ times to ensure that each image occurs in the test set only once and all images in
the whole dataset is either in the training set or in the test set at the same time. Finally, the
measures for each cross validation are averaged as the final metric of algorithm's performance.
Fig. \ref{fig3}(a) summarizes the steps of intra-CC.
The logic behind the typical intra-dataset-based evaluation is indeed
very suitable to test the performance of a learning-based CC model in the
presence of the multiple scenes with diverse illuminants and reflectances in a dataset \cite{gijsenij2011computational}.

Despite the significant advancement in the performance of intra-CC on several benchmark datasets
\cite{gijsenij2011computational,cheng2015effective,finlayson2013corrected,
bianco2014adaptive,shi2011illumination,forsyth1990novel,gijsenij2010generalized,
bianco2015color,cardei1999committee}, these methods always implicitly assume that
the distribution of the test data should be similar to the distribution of the training
data (e.g., both the training images and test images are captured by a fixed camera).
However, for practical CC applications this assumption may be readily
violated, since various trademarks of cameras possess quite diverse CSSs (e.g., NIKON and CANON in Fig. \ref{fig2}),
and there is even apparent difference of CSS among cameras produced by the same manufacture
(e.g., CANON in Fig. \ref{fig2}). Therefore, although the existing state-of-the-art algorithms
that are correctly trained can achieve high performance on intra-CC, they may also
suffer serious problem once being applied on inter-CC \cite{cheng2015effective,finlayson2013corrected,bianco2015color,funt2004estimating,cardei1999committee}.
\subsection{Problem formulation for inter-dataset-based CC}
Fig. \ref{fig3}(b) summarizes the general steps of the traditional inter-CC, we assume that we have trained a CC model $\Gamma$ on a dataset
that is rendered under certain CSS (we called it as CSS-1).
\begin{equation}
\begin{aligned}
I^{1}_{c}\!\approx \!\Gamma\{K(f^{1}_{c}(x))\}
\end{aligned}
\vspace{-0.3em}
\end{equation}
As indicated by Eqs (3) and (4), both the illuminant $I^{1}_{c}$ and the image $f^{1}_{c}(x)$ are mixed with the information of CSS-1.
Similarly, the color domain of feature statistics $K(f^{1}_{c}(x))$ extracted from $f^{1}_{c}(x)$ is also dependant on CSS-1.
Thus, based on these training data, the model $\Gamma$
just learns a fixed mapping that is only suitable to this fixed CSS-1. In other words, the model
$\Gamma$ will always learn a mapping that attempts to predict the illuminant under a color domain of
the corresponding CSS (e.g., a color domain rendered under CSS-1).

Then, to run the inter-CC, we need to test the CC model $\Gamma$ that
is trained on the color domain of CSS-1 on another dataset
that is collected by another camera with a distinct CSS (we called it as CSS-2).
\begin{equation}
\begin{aligned}
I^{2}_{c}\!\approx \!\Gamma\{K(f^{2}_{c}(x))\}
\end{aligned}
\vspace{-0.3em}
\end{equation}
Where $I^{2}_{c}$, $f^{2}_{c}(x)$, and $K(f^{2}_{c}(x))$ are rendered
under CSS-2. If we directly apply the CC model $\Gamma$
on the dataset that is rendered under CSS-2, the CC model $\Gamma$ is destined to suffer serious failure, which is explained as below.

While these learning-based CC approaches that are correctly trained
can deliver very competitive performance with intra-CC, the training phase
is undoubtedly relied on the illuminant ground truth supplied by the dataset \cite{gijsenij2011computational}.
As analyzed above, the color domains of the images and illuminants are clearly affected by the changes of CSS
(e.g., Fig. \ref{fig1}(b) and Fig. \ref{fig2}). Thus, both $I_{c}$ and $f_{c}(x)$ are color domain
dependent and the existing learning-based CC models can just learn a fixed model
that is only applicable to a fixed camera.

In other words, the existing learning-based
CC algorithms always train a model that tends to predict illuminant under a specific
color domain (e.g., a color domain rendered under specific CSS). In consequence, once
the trained CC model is tested on the images that are captured by another camera with
very different CSS, it is destined to suffer failure since they do not consider the effects of CSS during the training.

%Moreover, almost all of the existing learning-based algorithms just devote to discount the color bias introduced by the illuminant. However, the color of the image post corrected by those algorithms also relies on the CSS (Fig. 1 (b)). Although, there are several methods
%In contrast, our current work not only aims to improve the inter-dataset based CC performance, but also to color stabilize the resulting color corrected images for different CSS.
%\begin{figure}[t]
%\begin{center}
%\includegraphics[angle=90,width=0.5\textwidth]{fig4.eps}
%\end{center}
%\vspace{-1em}
%\caption{The overview of inter-dataset-based CC.}
%\label{fig4}
%\end{figure}
\begin{figure}[t]
\begin{center}
\includegraphics[angle=0,width=0.45\textwidth]{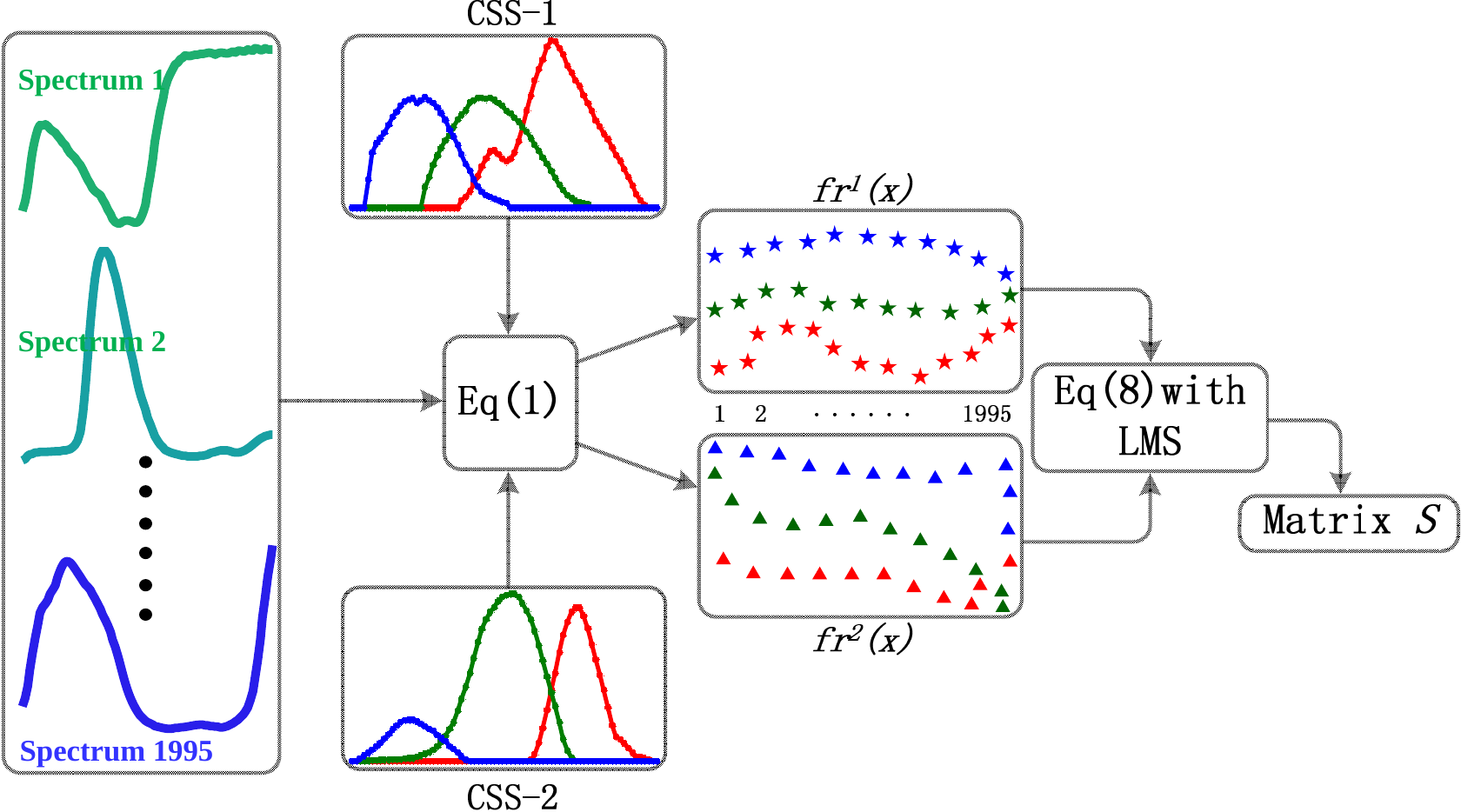}
\end{center}
\vspace{-1em}
\caption{The procedure of learning the transformation matrix \emph{S} between CSS-1 and CSS-2 based on the 1995 reflectance.}
\label{fig4}
\end{figure}
\section{Improving the inter-dataset-based CC}
In the above we have provided theoretical analysis about the effect that a CSS has on the
chromatic distribution of light source spectra
(Fig. \ref{fig2}) and the color appearance of images (Fig. \ref{fig1}(b)),
which further indicates that the existing CC models will perform poorly once being tested with inter-CC.

In this section, we propose a method to improve the performance of a CC
algorithm on inter-CC evaluation by taking into account the CSS of each camera.
Our strategy is straightforward. Specifically, in order to overcome the problem due to the difference between the two CSSs during the process of inter-CC, we first learn a sensor transformation that can express the mapping between the two given CSSs, and then we use this mapping to relate the CC model learned on one dataset with CSS-1 to another dataset with CSS-2. The improved framework for CC on inter-CC application is illustrated in Fig. \ref{fig5}.

Sensor transformation can be simply described by a $3\!\times\!3$ matrix
\cite{hubel1997matrix}. If $fr^{1}(x)$ denotes the response of CSS-1 to a certain reflectance, and $fr^{2}(x)$
denotes the response of CSS-2 to the same reflectance, we define
\begin{equation}
\begin{aligned}
fr^{2}(x)\!\!=\!\!S\!\!\cdot\!\! fr^{1}(x)
\end{aligned}
\end{equation}
where \emph{S} is a $3\!\times\!3$ matrix, which is simply learned by a least-mean-square (LMS)
training technique in this work. In order to learn the matrix \emph{S} based on Eq (8), we always use the 1995 reflectance spectras complied from the SFU hyperspectral dataset
\cite{barnard2002data} as the input to produce the responses $fr^{1}(x)$ and $fr^{2}(x)$ based on Eq (1), which produces the matrix size of $1995\!\times\!3$ for both $fr^{1}(x)$ and $fr^{2}(x)$. The procedure of learning \emph{S} is shown in Fig. \ref{fig4}.

For the inter-CC, this matrix $S$ is
utilized to transform both the image $f^{1}(x)$ and
illuminant $I^{1}_{c}$ rendered under CSS-1 into CSS-2. After this transformation, the CC model $\Gamma$ is then trained on this transformed data.
\begin{equation}
\begin{aligned}
S\!\cdot\!I^{1}_{c}\!\approx \!\Gamma\{K(S\!\!\cdot\!\!f^{1}_{c}(x))\}
\end{aligned}
\vspace{-0.3em}
\end{equation}
Finally, the trained model is tested on the images rendered under CSS-2 for the inter-CC,
i.e., to obtain the illuminant $I^{2}_{c}$ according to
\begin{equation}
\begin{aligned}
I^{2}_{c}\!\approx \!\Gamma\{K(f^{2}_{c}(x))\}
\end{aligned}
\vspace{-0.3em}
\end{equation}

It is worth to stress that learning this $3\!\times\!3$ matrix \emph{S} is very easy and only needs very few publicly available reflectance, which is normally far less than the number of images required to train an intra-CC model for each camera. Moreover, we have experimentally found that utilizing a same learned $3\!\times\!3$ matrix is enough to capture the transformation between two different CSSs for either Mondrian-like, hyperspectral, or real RGB dataset. Such attribute is very important since this indicates that the learned CSS adaptation matrix works independent of the scene. This makes our strategy widely applicable since we can use existing public surface reflectance samples (e.g., the 1995 reflectance spectra \cite{barnard2002data} used in this study) to learn a transformation matrix for each camera for later use, which can save much time and effort to prepare training set for new cameras (e.g., manually labeling numerous images under various environments and illuminants for each camera).
\begin{figure}[t]
\begin{center}
\includegraphics[angle=0,width=0.4\textwidth]{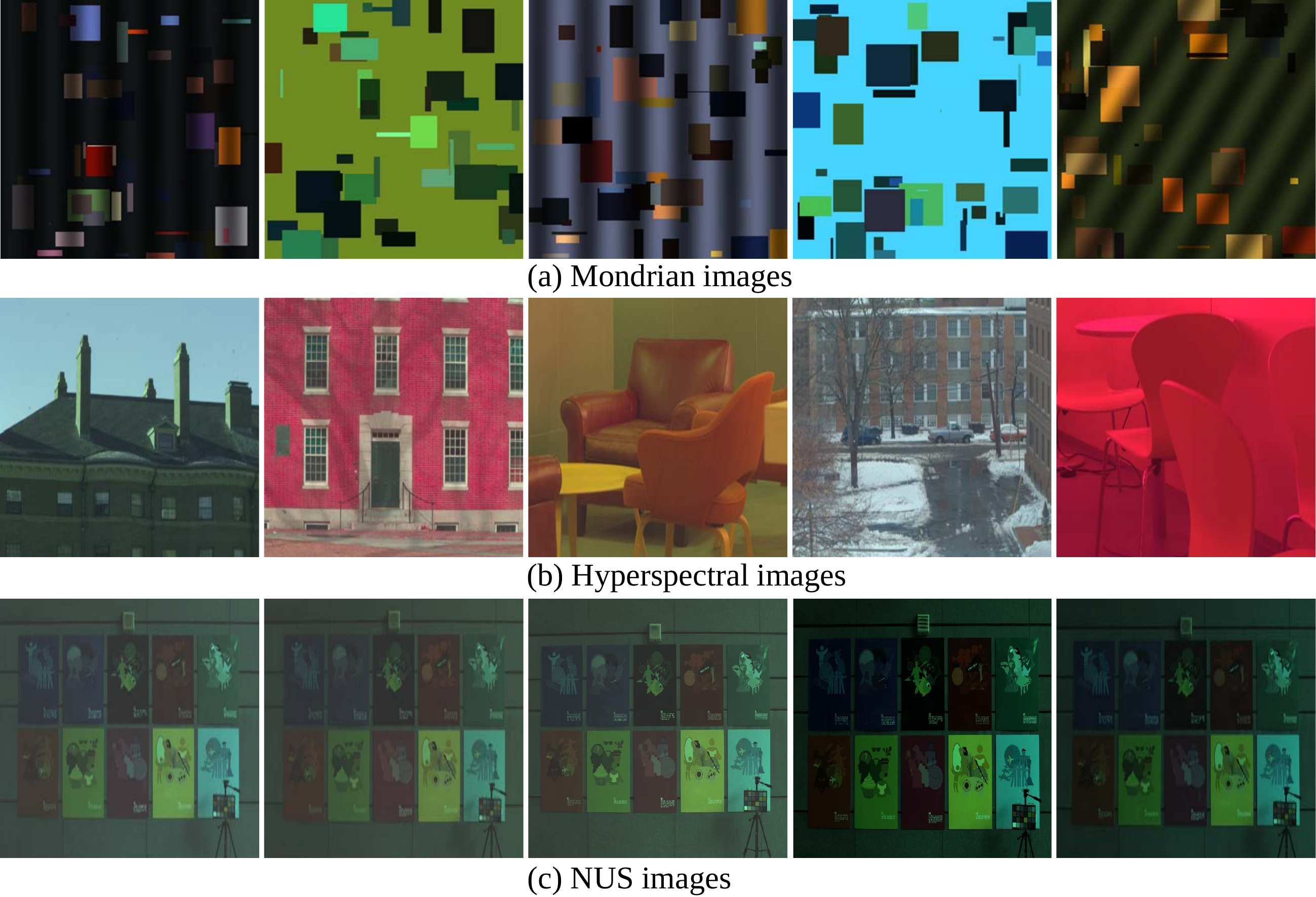}
\end{center}
\vspace{-1em}
\caption{Examples of (a) synthetic Mondrian-like, (b) hyperspectral, and (c) real camera captured images.}
\label{fig6}
\end{figure}
\begin{figure*}[t]
\begin{center}
\includegraphics[angle=0,width=0.9\textwidth]{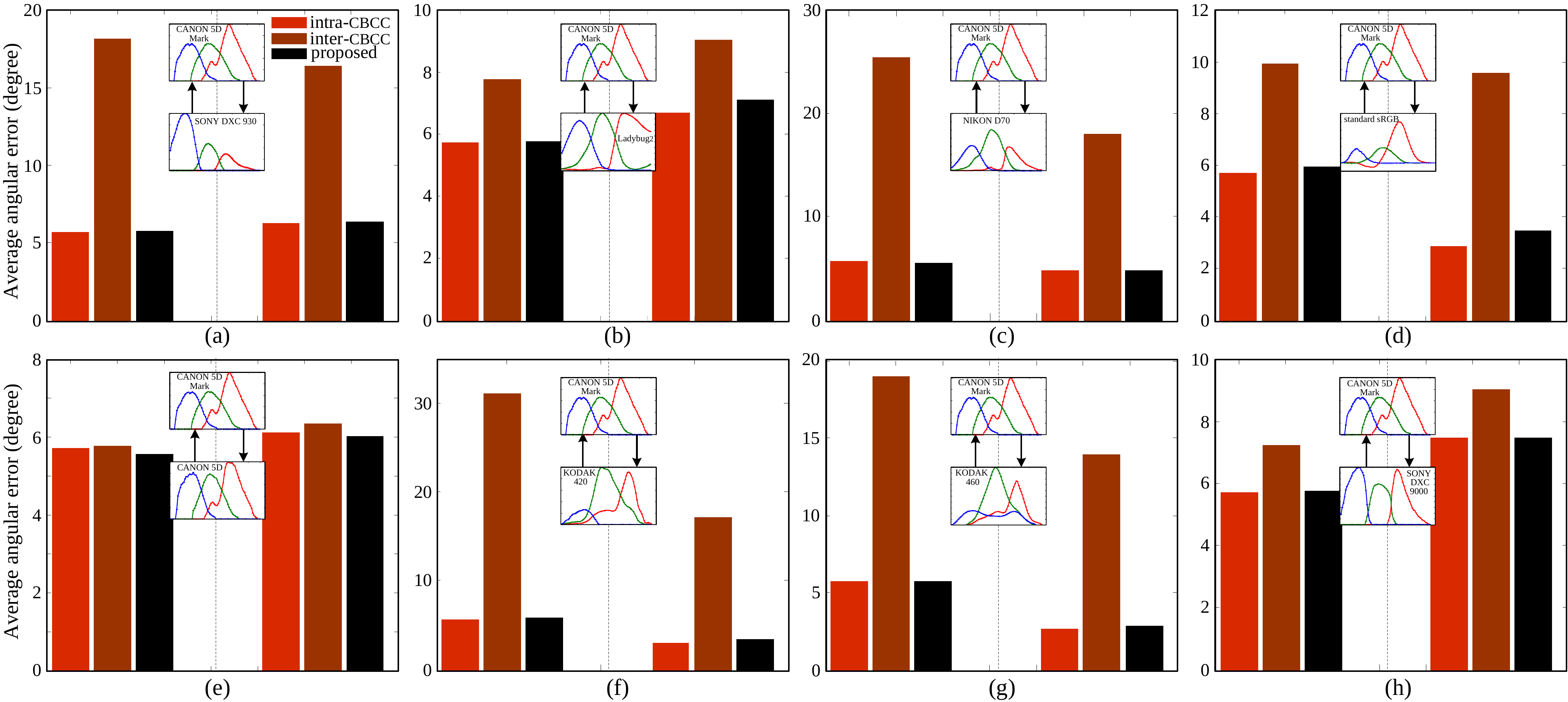}
\end{center}
\vspace{-1em}
\caption{The results of CBCC on the 14 Mondrian datasets. Each subgraph contains the results of intra-dataset-based CC (\emph{intra-CBCC}),
inter-dataset-based CC (\emph{inter-CBCC}) without considering the CSS effect, and the improved inter-dataset-based CC by taking the CSS into account (\emph{proposed}) on two datasets (the left three bars are on one dataset and the right on another dataset) with different CSSs (the CSS curves of the top and bottom panels are respectively for the left
and right datasets). In particular, the ``inter'' bar among the left three is the result of training on the right
dataset and testing on the left dataset, and vice versa. The notion to the ``intra'' bars are similar to that of the
``inter'' bars. Note that the intra-dataset-based (\emph{intra-CBCC}) results listed here were obtained using the three-fold
cross-validation, and the reported results were averaged over 100 random repetitions.}
\label{fig7}
\end{figure*}
\section{Implementations and experiments}
In this section, we first measure to what extent the effect of CSS has on the inter-CC performance with synthetic Mondrian and hyperspectral natural images. Then, the proposed method is comprehensively compared and validated on hyperspectral and real images coming from cameras. Finally, we exhibit several examples to show how a stable color representation of an image that is invariant to both the illuminant and CSS could be obtained with the proposed technique.

We select the early committe-based CC model (CBCC) \cite{cardei1999committee} as the
representative of the learning-based CC models to demonstrate the CSS effect.
Basically, there are mainly two considerations when choosing CBCC as the example.
On the one hand, CBCC can be almost taken as the earliest prototype of the regression-based CC models
and thus many of the existing state-of-the-art learning-based CC models can be included
into its framework \cite{cheng2015effective,finlayson2013corrected,gijsenij2011color,
funt2004estimating,
agarwal2006estimating,bianco2008improving,van2007using}. On the another hand, recent work \cite{cheng2015effective,
finlayson2013corrected,banic2015color} has indicated that with much simpler implementation,
CBCC can actually lead quite competitive performance in comparison to those more sophisticated learning-based CC
methods by incorporating with more effective features
(e.g., the color edge moments
used in CM \cite{finlayson2013corrected}). During our implementation, the outputs of the grey-world and
grey-edge models were integrated into the framework of CBCC to train the regression model in the LMS sense
\cite{finlayson2013corrected}. The ``cross terms'' employed in CM \cite{finlayson2013corrected}
that has been demonstrated to be very important for delivering the best CC result were also utilized in our implementation.

We also implement both the CM \cite{finlayson2013corrected} and spectral sharpening (SS) techniques \cite{finlayson1994spectral,vazquez2014spectral} to test their performance on inter-CC for comparison. CM is one of the state-of-the-art CC algorithms which improves the CC performance by learning a fixed matrix to correct the biased illuminant estimates of some low level based CC algorithms. In our implementation, we use 9-edge-moments-based CM. The aim we compare with the CM is to show if it is more convenient to correct the CSSs before applying CC algorithms (proposed) than to correct the illuminant estimates after applying CC algorithms (CM).

Spectral sharpening was originally proposed to sharpen the CSS such that each sharpened CSS has its spectral sensitivity concentrated as much as possible within a narrow band of wavelengths. Hypothetically, we expect that the sharpened CSSs could be more similar to each other and thus can improve the inter-CC performance of CBCC. Specifically, for a dataset CSS-1, we first convert the original sensors to its sharpened version via a $3\!\times\!3$ matrix multiplication and then learn the CBCC on the sharpened CSS-1. Then, given a second dataset CSS-2, we also perform spectral sharpening on the sensors of CSS-2 and finally test the performance of the learned CBCC of CSS-1 on the sharpened images of CSS-2.
\subsection{Validation on Mondrian-like images}
In this experiment we used a dataset containing 1995 spectra of reflectances and 102 spectra
of light sources compiled from several sources
\cite{barnard2002data} for the generation of Mondrian-like images. These spectra of reflectances and light sources were carefully collected under both
man-made and natural environments. To study the CSS effect, we used a recently published database
\cite{kawakami2013camera,zhao2009estimating}, which includes 12 sensors with various CSSs ranging
from common consumer level camera to industrial camera (several CSSs are shown in Fig. \ref{fig2}).
Moreover, the CIE color matching
function \cite{wyszecki1982color} and standard sRGB function
\cite{international1999multimedia} are also included as a kind of CSS during the test of CSS effect on CC.
\begin{figure*}[t]
\begin{center}
\includegraphics[angle=0,width=0.9\textwidth]{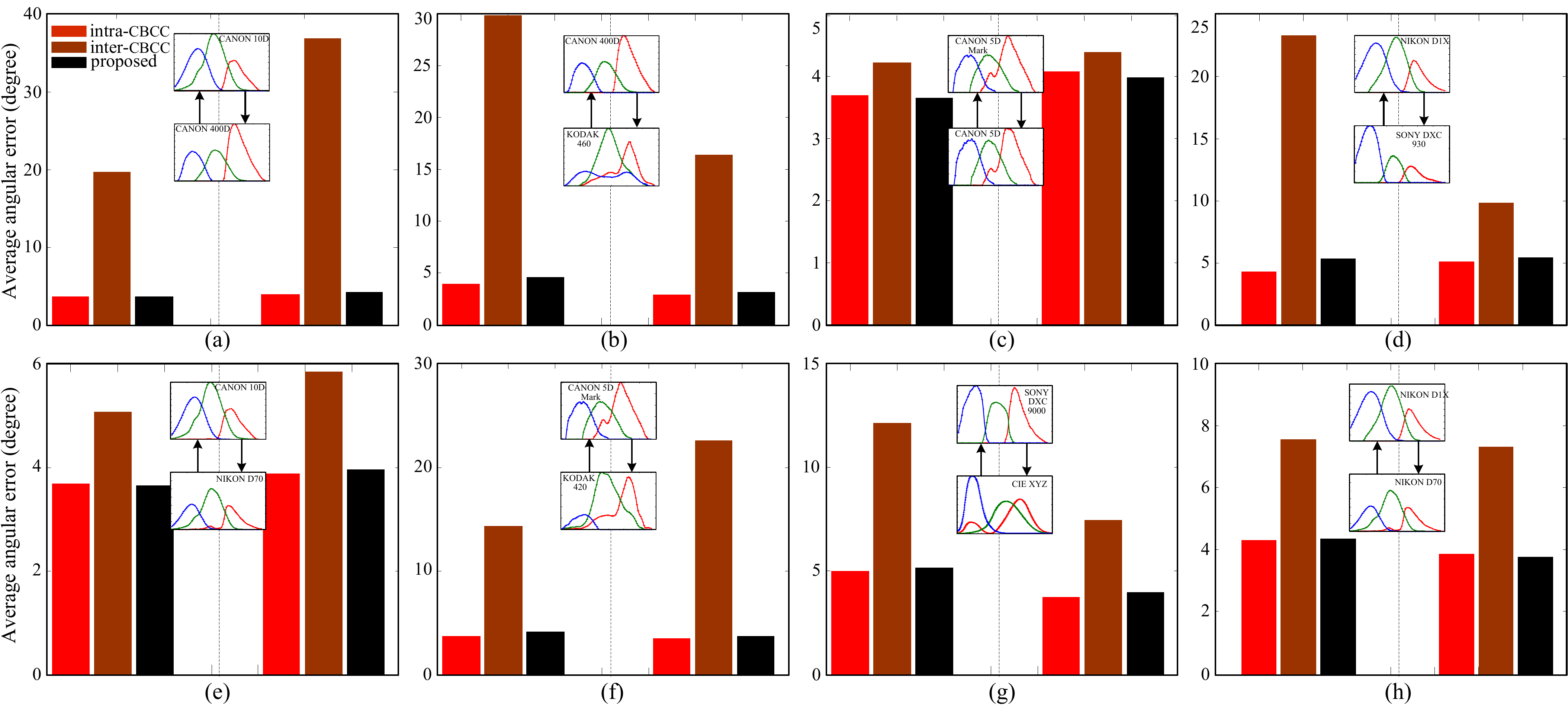}
\end{center}
\vspace{-1em}
\caption{The results of CBCC evaluated on the 14 hyperspectral datasets. The conventions are the same as those in Fig. \ref{fig7}.}
\label{fig8}
\end{figure*}

We generated the Mondrian-like datasets according to the model of image formation described by Eq (1). Specifically, the spectra of reflectance and illuminant \cite{barnard2002data} are first randomly selected and then integrated with CSS \cite{zhao2009estimating} over all the visible spectra to obtain R, G, B values. In practical computation, all spectrum curves are sampled and represented as vectors. Then, using these generated pixel colors, Mondrian-like images are created, and each image randomly contains up to tens of different surfaces, and hence many different transitions. The Gabor grating tuned brightness variations are also added in each Mondrian-like image for simulating as close as to the luminance of real-world images \cite{gijsenij2011color,
gao2015color,gijsenij2010generalized,barnard2002comparison}. Several examples are shown in Fig. \ref{fig6} (a).

We totally synthesized 14 Mondrian image datasets based on each CSS of 14 sensors
\cite{zhao2009estimating}. In each dataset, there are totally 510
Mondrian-like images (with a size of 300*300 pixels) composited by
random choices of illuminants and reflectances by using the aforementioned rules,
and hence can simulate the situations of the multiple scenes with diverse illuminants. Thus,
each dataset alone could be utilized to test the performance of intra-CC.

Moreover, among 14 Mondrian datasets, the scenes in each dataset are exactly same for all the cameras, but rendered by distinct CSSs. In other words, the only difference among the datasets is the CSS. Therefore, any pair of datasets arbitrarily selected from the Mondrian datasets could be used to accurately measure the CSS effect on the performance of inter-CC. For example, we trained CBCC on one Mondrian dataset that is specifically rendered under a certain CSS (e.g. CANON 5D Mark), and then tested the trained CBCC on another Mondrian dataset that is captured by a different CSS (e.g., NIKON D70).

Fig. \ref{fig7} summarizes the averaged performance of intra-dataset-based cross validation of CBCC on various Mondrian datasets (each dataset is named based on the used camera type). Angular error is usually used to test the accuracy of a CC algorithm \cite{gijsenij2011computational} by measuring the chromatic difference between the illuminant ground truth and the estimated illuminant by the CC algorithm.

Fig. \ref{fig7}
indicates that on all of the Mondrian datasets tested here, CBCC indeed obtains very good intra-CC performance. However, as discussed earlier, such evaluation is only suitable to fairly benchmark the performance of a CC method with the presence of multiple scenes with diverse illuminants. In contrast, for inter-dataset-based evaluation, CBCC suffers serious performance decreasing due to the CSS effects.

From the above evaluation on these 14 Mondrian datasets, we did not observe that there is CSS more effective than others for CC, and for the inter-dataset-based evaluation, we can observe based on the measurement of angular error in Fig. \ref{fig7} that the more similar the two CSSs are (e.g., the CSS between camera CANON 5D Mark and CANON 5D in Fig. \ref{fig7}(e)), the less impact of CSS is on the accuracy of either intra- or inter-dataset-based CBCC. In contrast, the more distinct the two CSSs are (e.g., the CANON 5D Mark and SONY DXC 930 in Fig. \ref{fig7}(a)),
the worse the performance of the original inter-dataset-based CBCC is. In short, the performance of the traditional inter-CC relies greatly on the similarity of CSSs among cameras. For example, when applying a model on a dataset that is trained on the images captured by a very distinct CSS (Fig. \ref{fig7}(c)),
very bad inter-CC performance is achieved. It is worth to note that for real camera captured images, the degradation caused by CSS is more complicated than Mondrian situation. We will further discuss this point in the following experiments using real camera captured images.
%Fig. \ref{fig6} also shows more examples including various types of CSSs produced by different manufactures, which represent various degrees of the CSS effect on the traditional inter-dataset-based performance of CBCC.
\subsection{Validation on Hyperspectral images}
While synthesizing Mondrian images according to Eq (1) is a pretty accurate first order model of image formation
\cite{finlayson2013corrected,wandell1987synthesis}, it is not able to model other reflective
effect (e.g., specular component) and thus may not reflect the real color image formation process.
Thus, we also utilized the hyperspectral natural images to measure
the CSS effect on the performance of inter-CC. For this
experiment we used a dataset containing 77 high quality hyperspectral images (with a size of 1392*1040 pixels)
acquired in real indoor and outdoor scenes \cite{chakrabarti2011statistics}.

Fig. \ref{fig6}(b) shows several examples of hyperspectral images employed in this experiments.
The subset of previous light source spectra dataset that contains 11 illuminants with both
Planckian and non-Planckian
\cite{barnard2002data}
and the 14 CSSs mentioned above were employed to produce 14
hyperspectral datasets. Each hyperspectral dataset totally contains 847 hyperspectral
natural images rendered under multiple illuminants for various investigations of the color
appearance of real-world scenes. Similar to the situation of Mondrian scenes, among the 14 hyperspectral
datasets, each one possesses identical distributions of reflectance and illuminant but exclusively rendered by distinct CSSs.
\begin{figure*}[t]
\begin{center}
\includegraphics[angle=0,width=0.9\textwidth]{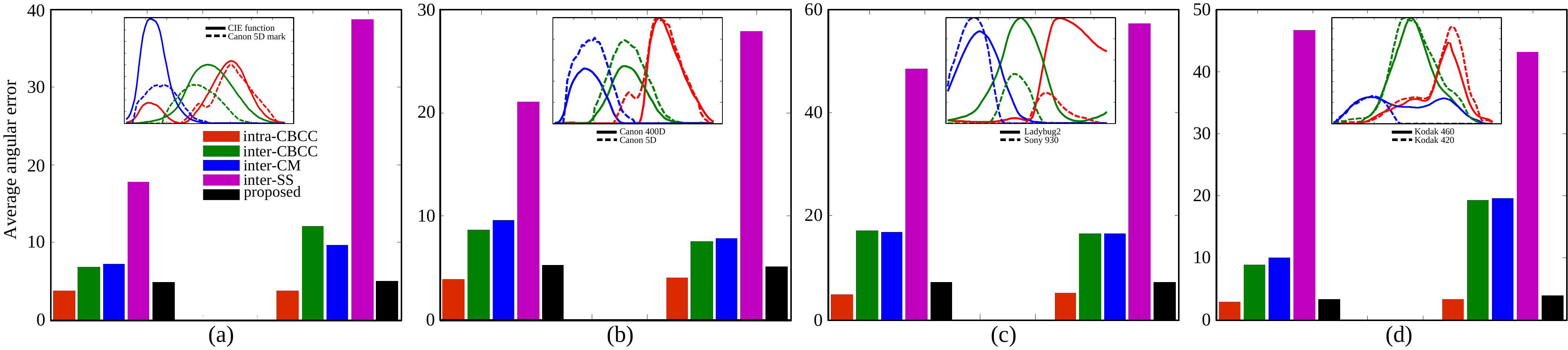}
\end{center}
\vspace{-1em}
\caption{The results of CBCC, CM, SS, and the proposed method evaluated on the hyperspectral datasets. The insets show the CSSs of two subsets.}
\label{fig9}
\end{figure*}
\begin{figure*}[t]
\begin{center}
\includegraphics[angle=0,width=0.9\textwidth]{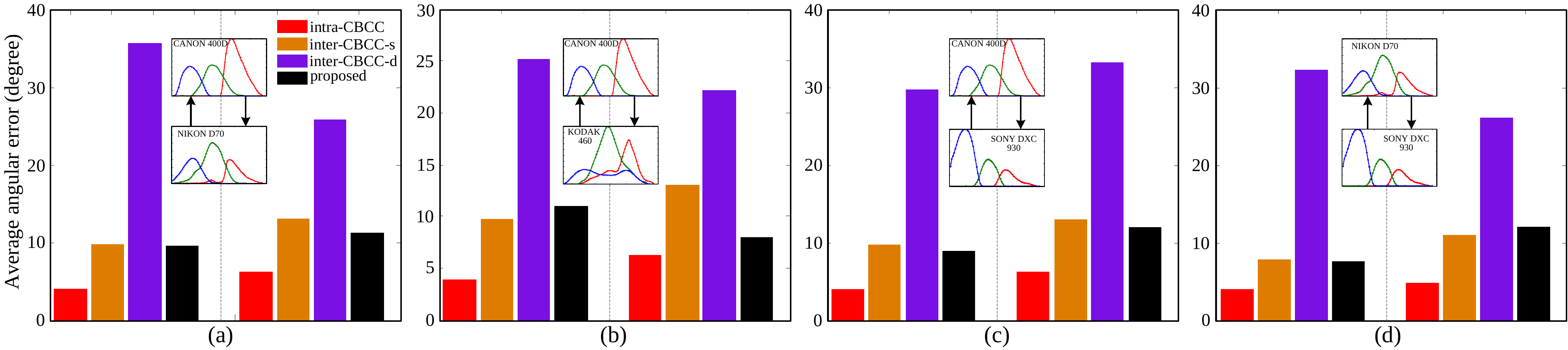}
\end{center}
\vspace{-1em}
\caption{The results of intra-based CBCC on the Mondrian dataset and the results of inter-based CBCC by training on the Mondrian set and testing on the hyperspectral dataset.
Each subgraph contains the results of intra-CC (\emph{intra-CBCC}), inter-CC with same CCSs (\emph{inter-CBCC-s}), inter-CC with distinct CCSs (\emph{inter-CBCC-d}), and
the improved inter-CC (\emph{proposed}).}
\label{fig10}
\end{figure*}
Fig. \ref{fig8} shows the results of CBCC on both intra- and inter-dataset-based evaluation. The observations on these results are quite consistent with those obtained in the previous experiments with synthetic Mondrian datasets shown in Fig. \ref{fig7}.
As a comparison to the experiments shown in Fig. \ref{fig8}, Fig. \ref{fig9} shows more results with the methods of CM and SS for inter-dataset-based evaluation. Similar to CBCC, the-state-of-the-art CM and early SS also suffer the drastic decreasing of performance on inter-dataset-based evaluation. The reason why CM fails to achieve a good performance for inter-dataset-based evaluation is that similar to CBCC, CM just learns a fixed matrix for specific camera \cite{finlayson2013corrected}. In other words, CM is not able to adapt to other cameras. For SS, although we assume that the sharpened CSSs are more similar to each other and thus SS could improve the performance of inter-dataset-based CBCC, the actual performance is surprisingly poor since we observed that the sharpened CSSs include many negative values, which may finally result in the large angular error for inter-dataset-based CC. This problem may be alleviated when using technique of spectral sharpening with positivity \cite{drew2000spectral}.
\subsection{Improving the inter-dataset-based CC}
In the above sections we have provided a systematic analysis of the CSS effect on the color appearance of images
(Fig. \ref{fig2}) and the chromatic distribution of light source spectra (Fig. \ref{fig1}(b)),
and shown how a CSS can significantly affect the performance of a learning-based CC algorithm on inter-dataset-based evaluation
(Fig. \ref{fig7}, Fig. \ref{fig8}, and Fig. \ref{fig9}).
In this section, we will show the improved performance of inter-CC by the proposed method.
The improved performance of CBCC on both Mondrian and hyperspectral datasets are shown in
Fig. \ref{fig7}, Fig. \ref{fig8}, and Fig. \ref{fig9}, respectively. We can observe that the performance of CBCC on
inter-dataset-based evaluation is greatly boosted by the proposed strategy that includes
the CSS effect during the model training. Surprisingly, by employing the proposed strategy,
the improved performance of CBCC on inter-dataset-based evaluation almost reaches the same level as that of the CBCC on intra-dataset-based evaluation.
\begin{figure*}[t]
\begin{center}
\includegraphics[angle=0,width=0.9\textwidth]{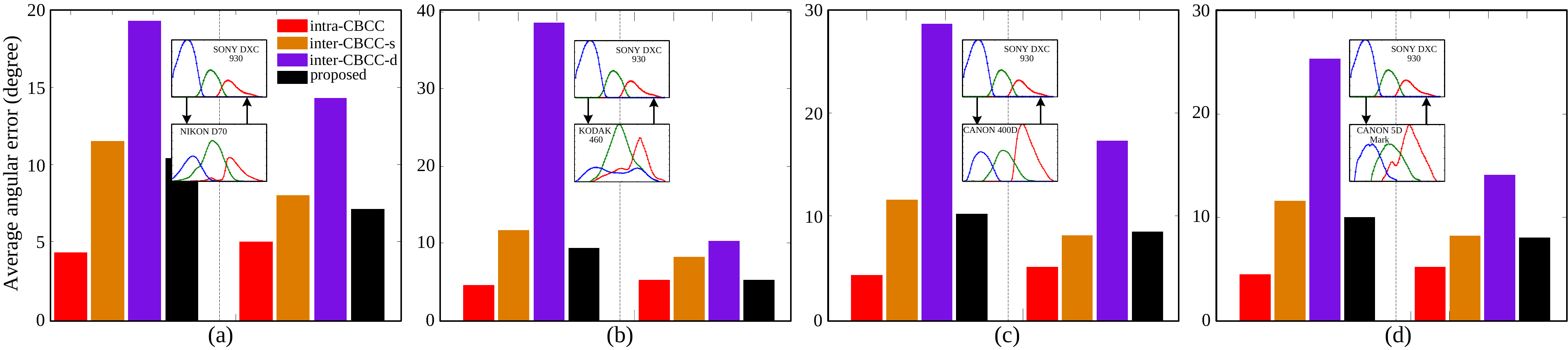}
\end{center}
\vspace{-1em}
\caption{The results of intra-based CBCC on the hyperspectral dataset and the results of inter-based CBCC by training on the hyperspectral set and testing on the real SFU lab dataset. See similar notions to the bars as indicated in Fig. \ref{fig10}.}
\label{fig11}
\end{figure*}
\begin{figure*}[t]
\begin{center}
\includegraphics[angle=0,width=0.9\textwidth]{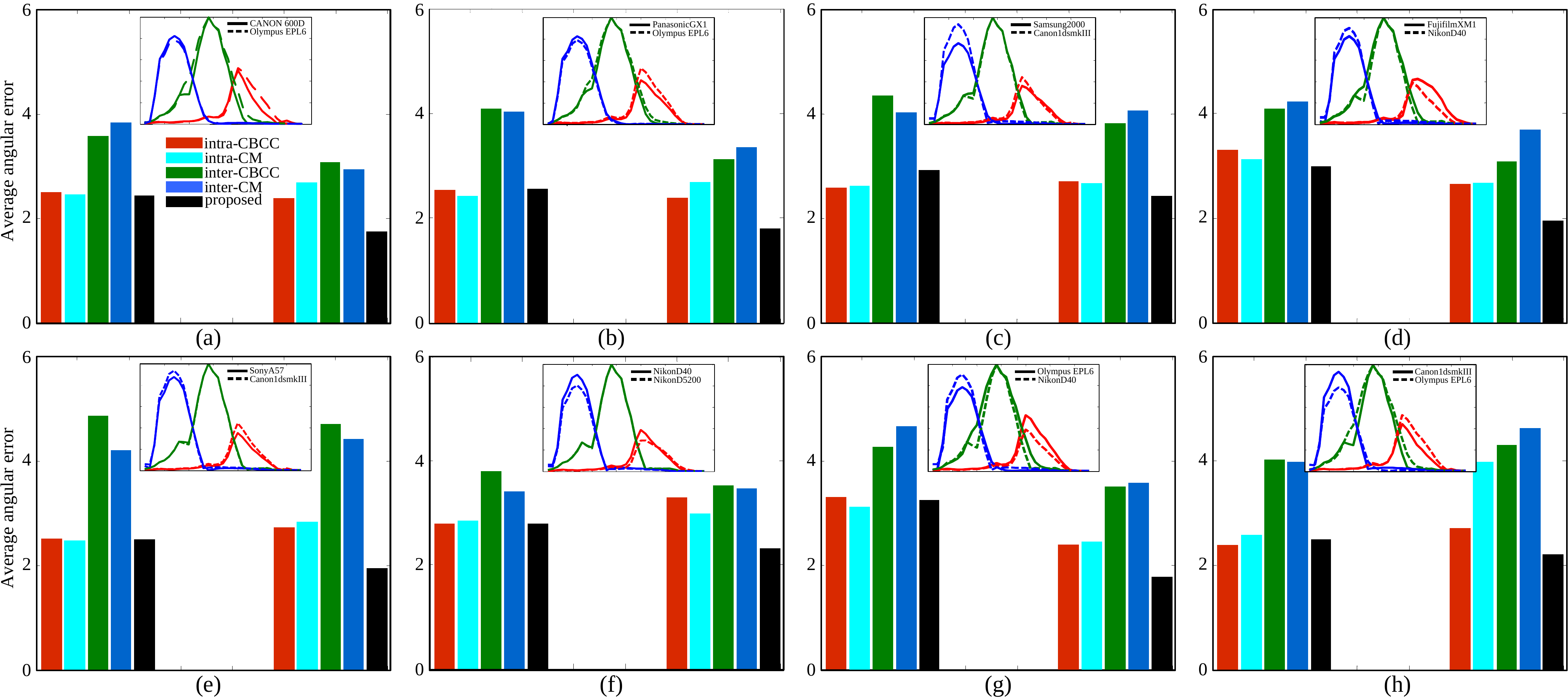}
\end{center}
\vspace{-1em}
\caption{The results of CBCC, CM, and proposed method evaluated on the NUS dataset \cite{cheng2014illuminant}.}
\label{fig12}
\end{figure*}
\subsection{More general situations}
We have shown how to improve the inter-CC by the proposed method on datasets that have exactly the same reflectance distributions but with distinct CSSs (Fig. \ref{fig7}, Fig. \ref{fig8}, and Fig. \ref{fig9}).
For practical application, however, the general situation is that the training set
not only has distinct CSSs but also possesses with different reflectance distributions compared to
the test set (e.g., the training dataset is captured under indoor environment, but the test dataset
is collected under outdoor natural environment). In this section, we measure the CSS effect on
such a more general situation.

Specifically, we trained the CBCC model on the Mondrian-like dataset and then tested it
on the hyeprspectral natural dataset. We also trained the CBCC on the hyeprspectral dataset and
tested it on a real RGB dataset
\cite{barnard2002data,barnard2002comparison1}
(SFU lab dataset) that is captured by SONY DXC 930. In these experiments, we evaluated the CBCC under
four situations including intra-dataset-based CC (intra-CBCC), inter-dataset-based CC with same
CSSs (labeled as inter-CBCC-s), inter-dataset-based CC with distinct CSSs (labeled as inter-CBCC-d),
and finally the improved model (proposed). Fig. \ref{fig10} shows the results of intra-based CBCC on the Mondrian dataset and the results of inter-based CBCC by training on the Mondrian set and testing on the hyperspectral dataset. Similarly, Fig. \ref{fig11} shows the results of intra-based CBCC on the hyperspectral dataset and the results of inter-based CBCC by training on the hyperspectral set and testing on the real SFU lab dataset.
%All these situations show the effect of various examples of CSS on the inter-CC.

In general, the experimental results for each situation behave like this: as is expected,
CBCC obtains very good CC performance of intra-dataset-based evaluation on all the datasets
rendered under any CSSs, illuminants and scenes. For the situation of inter-CC,
intuitively, the greater the difference exists between the training and the test datasets, the
worse the performance of CBCC obtains. For example, the performance of inter-based CBCC with
same CSSs (inter-CBCC-s) is worse than its performance on the intra-CC (intra-CBCC), since the two
datasets utilized for training and testing have different reflectance distributions (e.g., the scene
structure in hyperspectral dataset is very different from the scene structure in SFU lab dataset).
The worst performance of CBCC is obtained for the inter-CC with distinct CSSs (inter-CBCC-d)
due to that there is not only the huge difference of reflectance distributions but
also the huge difference of CCSs between the two datasets. This experiment once again demonstrates the adverse effect of CSS on the performance of a learning-based CC algorithm for the inter-CC application.

It should be noted that the improved performance of CBCC on inter-CC with different reflectance (inter-CBCC-s)
does not arrive at the same performance level of the intra-CC as shown in
Fig. \ref{fig7}, Fig. \ref{fig8}, and Fig. \ref{fig9}, which is mainly due to the huge difference of
reflectance distribution between the two datasets evaluated here (e.g., the difference
between the natural scene in hyperspectral dataset and the laboratory scene in SFU lab dataset).
Nevertheless, the performance of CBCC on inter-CC with different CSSs (inter-CBCC-d) is greatly improved
by the proposed strategy.
\subsection{Validation on real camera captured images}
All the above experiments used the synthetic images or multispectral ones, which ignore all the non-linearities that occur in digital camera pipelines before the RAW image is saved. To validate the applicability of the proposed method in a real-world scenario, we finally tested our proposed method on NUS dataset \cite{cheng2014illuminant}, which includes 1853 high quality linear images taken by 9 different cameras in real environment. In addition, this dataset is composed of images of the same scene with the different cameras, such that the scene and illumination is the same for all the 9 cameras. This makes the dataset suitable to investigate the CSS effect on inter-CC under real camera situation. One example of a same scene taken by 5 difference cameras is shown in Fig. \ref{fig6} (c). For unbiased evaluation, during experiments we masked out the color checker patch which is originally used to compute the illuminant ground truth of each image. Since SS performs very poorly in previous experiments (Fig. \ref{fig9}), we only compared the improved results of CBCC with CM.

Similar to what we have reported on the synthesized Mondrian-like and hyperspectral images, Fig. \ref{fig12} shows the results on several NUS subsets captured with various cameras. As expected, CBCC and CM perform very well on intra-dataset-based evalutation. However, once being applied in inter-CC setup, the performance of both methods is clearly degraded due to the effect of CSS, and in general, the larger the difference between the CSSs is, the worse the inter-CC performs for CBCC and CM.

It should be noted that all the cameras utilized for capturing the NUS dataset have very similar CSSs since they are specifically chosen to satisfy the so-called Luther condition for CSS approximation \cite{prasad2013quick}. Hence, the results also indicate that although such a small difference exists among different CSSs (cameras), which indeed leads to the drastic performance decreasing for the existing learning-based CC models on inter-dataset-based evaluation. Thus, these experiments on the dataset from real cameras further ground our theoretical analysis in subsection \uppercase\expandafter{\romannumeral2} (C). In short, in order to develop a CC algorithm with good generalization ability, we need to reasonably consider the CSS effect.
\begin{figure}[t]
\begin{center}
\includegraphics[angle=0,width=0.3\textwidth]{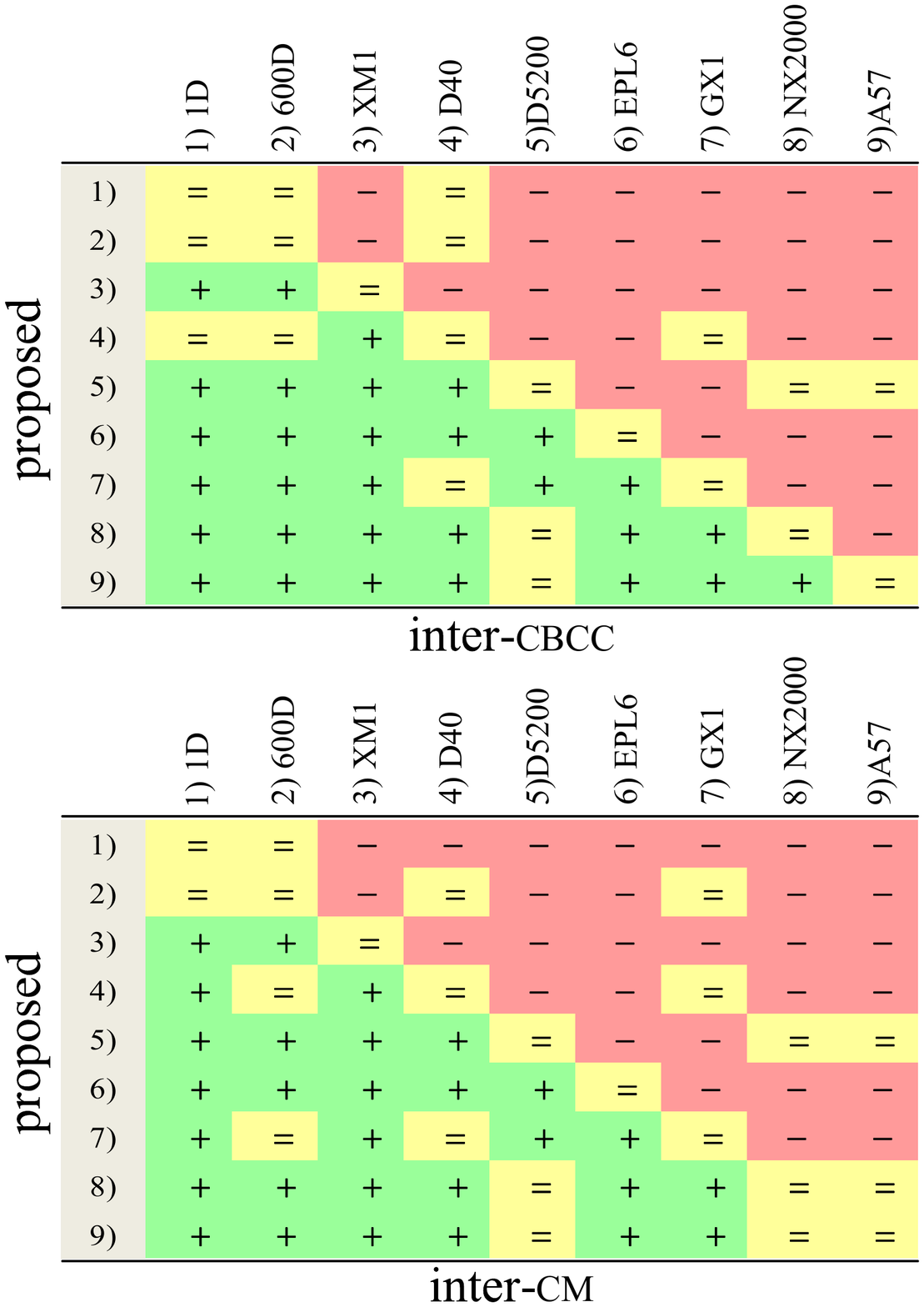}
\end{center}
\vspace{-1em}
\caption{WST test between the proposed method and the inter-CBCC (top), and inter-CM (bottom) in the NUS dataset.}
\label{fig13}
\end{figure}
%We noted that in Fig. \ref{fig11}(e), the two CSSs is quite similar but the inter-dataset-based performance of CBCC and CM is also decreased much. We have checked the images collected in those two CSSs and found that there are some part of images, which are quite different although the images were collected in the same location by using those two CSSs (e.g., Fig. \ref{fig14}). Thus, we attribute this to the fact that the feature distribution of the training dataset is partly different from that of the testing dataset, which also leads to the further decrease of inter-dataset-based CC as indicated in previous experiments (e.g., Fig. \ref{fig9} and Fig. \ref{fig10}). Thus, there are mixture factors of scene content and CSS that result in the performance decreasing of the inter-CC \cite{joze2014exemplar,gao2015color,wu2015edge}. In spite of this, our proposed method always obtains better performance than CBCC and CM on inter-dataset-based evaluation (e.g., Fig. \ref{fig11}).

We noticed that with the quite similar CSSs between two cameras for the NUS dataset, the inter-CC performance by CBCC and CM are still greatly degraded in comparison to the intra-CC performance (e.g., Fig. \ref{fig12}(e)). Before definite reason can be found to explain this phenomenon, we speculate that based on Eq (1), besides the CSS, there are other mixed factors determining the color appearance of a captured image, and consequently, the inter-CC performance should also be influenced at least by the distributions of illuminant and surface reflectance. Let us take Fig. \ref{fig12}(a) as an example. If we make a mutual change of the training and test sets, the performance of inter-CBCC varies markedly, as indicated by the green bars on the left and right parts of Fig. \ref{fig12}(a). Similar observations can also be found for the inter-CC evaluations on the Mondrian and hyperspectral datasets (Figs. \ref{fig7}-\ref{fig11}). This means that besides the difference between the CSSs, the inter-CBCC performance is affected determinately by the difference between the training and test sets. This may partially explain the poor inter-CC performance on two datasets even with the visual similarity of CCS. In fact, such phenomenon further emphasizes the difficulty of color constancy in real applications, where the difference between the inherent features of training and test images is uncontrollable.

In order to further determine whether the proposed method significantly improves the performance of CBCC and CM on inter-CC, we further utilized the Wilcoxon Signed-Rank Test (WST) to measure the performance difference between the inter-CC (e.g., inter-CBCC and inter-CM) and our method. WST has been
recommended as an valuable tool for performance
evaluation \cite{gijsenij2011computational,gao2015color,hordley2006reevaluation}. In this study, given any
two different algorithms, the WST was run on their angular
error distributions on the whole dataset, and its result was
used to conclude that at a specific (e.g., 95\%) confidence level, the angular error of one algorithm is often
lower or higher than that of another algorithm, or there is
no significant difference between them.

Fig. \ref{fig13} reports the results of the statistical significance test using WST on all cameras. A sign ($+$) at location (i, j) indicates that the
average angular error of method i is significantly lower than
that of method j at the 95\% confidence level, and a
sign ($-$) at (i, j) indicates the opposite situation. A sign ($=$)
means that the average angular errors of the two methods have no
significant difference. Fig. \ref{fig13} indicates that the performance of the
proposed method exhibits significant improvement over both CBCC and CM in the most situations of inter-CC evaluation.
%\begin{figure}[t]
%\begin{center}
%\includegraphics[angle=0,width=0.3\textwidth]{fig14.eps}
%\end{center}
%\vspace{-1em}
%\caption{Two examples of a same scene captured by Camera Sony A57 and Cannon1D, which show quite distinct color appearance although the visual similarity of their CSSs.}
%\label{fig14}
%\end{figure}
\subsection{Stable color representation of images}
In previous sections, we have shown the improvement of inter-CC by incorporating CSS
information. Here, we show examples of actual improvement on image appearance by
further considering the CSS after CC. Even the two images in Fig. \ref{fig1}(b)
have been corrected by CC, they still exhibit obviously different color appearance due to the
difference between CSSs. In order to eliminate the chromatic difference induced by CSSs, the original color
biased images taken by the two CSSs were respectively corrected by a CC algorithm (here by the CC method of grey-edge \cite{van2007edge}), then a transformation
by the learned matrix was further used to adapt the corrected image rendered under one CSS (e.g., KODAK DCS 460)
to another CSS (e.g., CANON 5D Mark 2).
\begin{figure*}[t]
\begin{center}
\includegraphics[angle=0,width=0.9\textwidth]{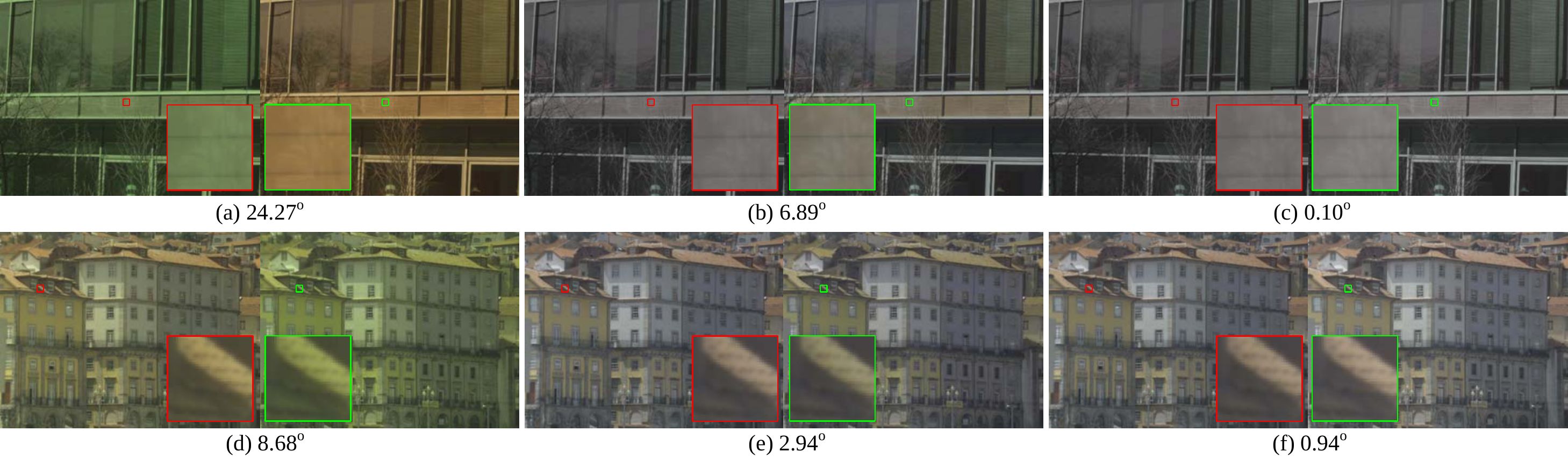}
\end{center}
\vspace{-1em}
\caption{Results on two pairs of images from \cite{chakrabarti2011statistics,nascimento2002statistics}. The first column lists two pairs of scenes rendered under different illuminants and CSSs. The second column shows the images after removing the the effect of illuminant. The last column lists the results after further discounting the CSS effect.}
\label{fig14}
\end{figure*}

Fig. \ref{fig14} shows examples selected from the two hyperspectral datasets
 \cite{chakrabarti2011statistics,nascimento2002statistics}. It is clear that the color cast on the image pairs due to the different CSSs and illuminants are well removed after applying a CC algorithm and
then eliminating the CSS effect. For example, Fig. \ref{fig14}(a) shows obvious chromatic difference
between the two images since they are rendered under distinct illuminants and CSSs.
By applying CC on these two images, although the color bias induced by the illuminants are removed
(Fig. \ref{fig14}(b)), these two images still present significant chromatic difference since
they are rendered under two different CSSs. However, the chromatic difference of these
two images is further eliminated after removing the CSS effect by the proposed method (Fig. \ref{fig14}(c)).

It should be noted that several CC methods have been proposed to impose stabilization on the resulting color corrected images through specific constraints. For example, in \cite{sapiro1999color} Sapiro uses the probabilistic Hough transform to introduce the physical constraints that the corrected solution should accomplish; in \cite{finlayson2001color1} Finlayson et al. select the illuminant from a predefined subset that better correlates with a set of given reflectances, and in \cite{vazquez2012color} Vazquez-Corral et al. select the illuminant that maximizes the number of psychophysically based focal colors presented in the corrected scene. However, the main aim of those constraints is to improve the accuracy of illuminant estimation instead of explicitly removing the color bias triggered by CSS. In other words, the results of these algorithms is still CSS dependent.

\section{Discussion and Conclusion}
In this paper we demonstrated that CCS is a quite important factor that results in the serious degeneration of learning-based CC algorithms in inter-dataset-based application due to that both the illuminant and image are actually color domain dependent (camera dependent). It means that the existing CC algorithms only build very limited color correction models that are only suitable for a specific camera. Thus, once applying the trained CC model on the images collected by a camera with distinct CSS, the model will undoubtedly suffer serious problem. We have grounded our theoretical analysis on experiments with various datasets.

We also proposed a simple yet effective way to incorporate the CSS information into the training process through an adapted matrix, which builds a mapping between two CSSs. As a direct consequence of our strategy, we can model CSS effect when building a CC model. Comprehensive experiments on synthetic, hyperspectral, and real camera captured datasets have shown how the embedded CSS information can greatly improve the performance of learning-based CC algorithms on inter-dataset-based evaluation. Many existing state-of-the-art learning-based CC methods will benefit from our proposed technique to improve their generalization ability across multiple cameras.

As a practical application, we have also shown examples to demonstrate how a stable color representation of scene across the changes of illuminants and CSSs could be obtained by the proposed method. This is an important step towards the normalization of color appearance under different acquisitions with different devices.

The results presented in this work prove very useful for many color-based applications. Imagine we are facing a surveillance problem using a large scale camera network and we want to build a CC model for this camera network. For traditional implementation, we have to first capture a large group of images under diverse illuminants for each camera, and then train a CC model separately for each camera utilizing the corresponding manually labeled images and illuminants with this camera. This will not be practical in this large scale camera network, since manually labeling many images under various illuminants for each camera is very expensive and time-consuming. However, by employing the solution proposed in this work, it is enough to train a CC algorithm for any camera with the only information of different CSSs by just employing the images under various illuminants captured by only one camera, since we can adapt a CC algorithm between any two cameras by taking the CSSs into account.

Moreover, the recovered images by the proposed strategy
(Fig. \ref{fig14}) are almost the truly color constant images since they are independent of both CSSs and illuminant, which would be very useful for further computer vision tasks that need accurate color characterization of object materials invariant to specific device characteristics
(e.g., intrinsic image decomposition
\cite{barrowrecovering,
serra2014photometry}). Furthermore, we believe that for some widely used sensors, such stable corrections could even be incorporated in the standard libraries for wide use in any application that relies on accurate matching of surface appearances across different images.

In this work, we assume that the CSS utilized to calculate the adapted matrix
between two CSSs is already known. This is practically
applicable since such information is mostly publicly available for many camera sensors
(e.g., http://www.dxomark.com/) \cite{serra2014photometry}, or we can estimate it using existing CSS estimation algorithms
\cite{kawakami2013camera,zhao2009estimating,jiang2013space,barnard2002camera,prasad2013quick}.

Another point that deserves a brief comment is that all the analysis in this work is true when we are dealing with RAW images. In case that the images are rendered to sRGB and post-processed with some nonlinear transformations built in the camera, the inter-dataset-based CC problem becomes more difficult since we will need to learn a transformation for each particular set of camera parameters (e.g., style look).

This work also initiates an open question that how to build a general CC model that could be applied for any cameras and situations. Our proposed method is good at dealing with the failure of inter-CC situation caused by different CSSs. However, during the experiments we also observed that there are other factors (e.g., scene content) which also affect the performance of inter-CC. Thus, it is very important to employ other useful information during building a CC model with higher efficiency and plasticity, which is our future work.
\section*{Acknowledgment}
We would like to thank Dr. Dilip Kumar Prasad and Michael S. Brown for providing their CSSs of NUS dataset and Dr. Javier Vazquez-Corral for sharing the code of spectral sharpening. We would also thank the anonymous reviewer's critical comments on the original version of our manuscript.

\ifCLASSOPTIONcaptionsoff
  \newpage
\fi



%\bibliographystyle{IEEEtran}
%\bibliography{doublePMAI}

\bibliographystyle{IEEEtran}
\bibliography{CSS}

%\bibliographystyle{IEEEtran}

%\bibliography{doublePMAI}

%\begin{thebibliography}{1}
%\end{thebibliography}


%%Biography text here.
%\end{IEEEbiography}

\begin{IEEEbiographynophoto}{Shao-Bing Gao}
received his M.Sc. degree in Biomedical engineering from University of Electronic Science and Technology
of China (UESTC) in 2013, where he is currently pursuing his Ph.D. degree
in Biomedical engineering. His research interests include visual
mechanism modeling and image processing.
\end{IEEEbiographynophoto}
%
%%\begin{IEEEbiographynophoto}{Xian-Shi Zhang}
%%received a B.Sc. and M.Sc. degree in Pattern recognition and intelligence system
%%from UESTC in 2005 and 2008, where he is currently pursuing his Ph.D.
%%degree in Biomedical engineering. His research interests include
%%visual mechanism modeling and image processing.
%%\end{IEEEbiographynophoto}
%
\begin{IEEEbiographynophoto}{Ming Zhang}
received his B.Sc. degree in Biomedical
engineering from UESTC in 2015, where he is currently pursuing his Master
degree in Biomedical engineering. His research interest is image processing.
\end{IEEEbiographynophoto}
\begin{IEEEbiographynophoto}{Chao-Yi Li}
graduated from the Chinese Medical University in 1956
and Fudan University in 1961. He became an academician of the
Chinese Academy of Sciences in 1999. He is currently a professor
of UESTC, and
professor of Shanghai Institutes for Biological Sciences. His research
interest is visual neurophysiology.
\end{IEEEbiographynophoto}
\begin{IEEEbiographynophoto}{Yong-Jie Li}
received his Ph.D. degree in biomedical engineering
from UESTC in 2004. He is now a professor of the Key Lab for
Neuroinformation of Ministry of Education, School of Life Science
and Technolgy, UESTC, China. His research interests include visual
mechanism modeling and image processing.
\end{IEEEbiographynophoto}

%\newpage



%\vfill


\end{document}